\documentclass[letterpaper, 10 pt, conference]{ieeeconf}  

\IEEEoverridecommandlockouts                              

\usepackage{amsmath} 
\usepackage{amssymb}  
\usepackage{graphicx}
\title{\LARGE \bf
DeltaSeek: Toward Active Perception in Evolving Construction Environments
}

\author{Sanjay Acharjee$^{1}$ and Md Nazmus Sakib, Ph.D.$^{2}$
\thanks{$^{1}$Department of Civil Engineering,
            University of Texas at Arlington, TX 76019, USA.
            {\tt\small sanjay.acharjee@uta.edu}}%
\thanks{$^{2}$Department of Civil Engineering,
            University of Texas at Arlington, TX 76019, USA.
            {\tt\small mdnazmus.sakib@uta.edu}}%
}

\begin{document}

\maketitle
\thispagestyle{empty}
\pagestyle{empty}

\begin{abstract}

Construction environments evolve continuously, causing large geometric changes that degrade static mapping and registration performance. This necessitates active perception, where robots deliberately select sensing configurations to resolve the environment's current state. We present DeltaSeek, an initial framework toward active perception in evolving built environments. While our broader objective is a system that reasons about where, how, and when to observe, this paper addresses a critical prerequisite: how a robot's sensing embodiment constrains the observations it can acquire. We formalize an embodiment's permissible observation set and evaluate with a Husky A300 equipped with a UR5e on an IFC-derived benchmark under chassis-mounted and wrist-mounted RGB-D configurations, scoring observations by geometric visibility and effort by drivable distance. In a room-scale scene with eight controlled changes spanning four observability conditions, exhaustive evaluation over 240 permissible base poses and five arm postures shows that two changes admit no chassis viewpoint whatsoever, while the wrist camera resolves both. For changes observed by both embodiments, the median base travel is $6.0$~m for the wrist camera and $15.2$~m for the chassis camera. These results distinguish sensing limitations from acquisition costs, clarifying whether an observation is impossible or simply requires more travel.

Code and dataset: https://github.com/iSET-LAB/deltaseek

\end{abstract}

\begin{figure*}[!t]
    \centering
    \includegraphics[width=\textwidth]{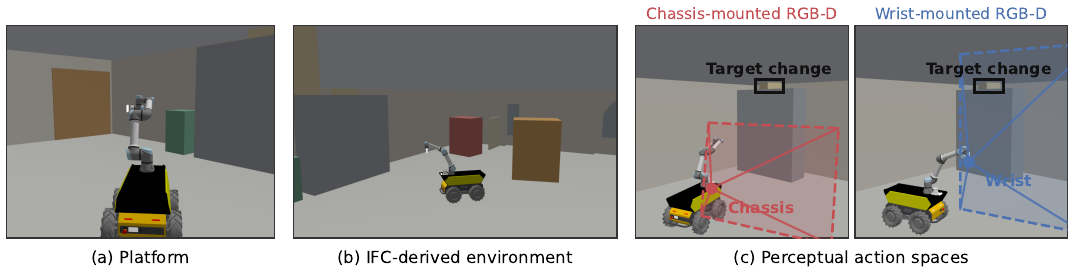}
    \caption{Simulation study. (a) Husky A300--UR5e platform with identical chassis- and wrist-mounted RGB-D cameras. (b) IFC-derived Hall B with eight injected changes. (c) Evaluated frusta for the same target, which lies outside the chassis-camera frustum but inside the wrist-camera frustum. Each view uses the base yaw that directs its sensor toward the target.}
    \label{fig:overview}
\end{figure*}
\section{INTRODUCTION}

Construction sites are dynamic. Between two separate visits, walls are erected, services are hung, pallets are delivered and consumed, and formwork is struck. The changes are precisely what a verification system cares about~\cite{sun2025nothing}. Currently, mobile robots deployed for site scanning rely predominantly on passive accumulation: a coverage route is driven, point clouds or images are collected, registered to the model, and differenced offline~\cite{bosche2010automated, turkan2012automated, golparvar2015automated}. Registration depends on correspondence between what was observed before and what is observed now, and large change is exactly the regime in which that correspondence becomes sparse: the information any autonomous robot most requires is the information its pipeline is least equipped to acquire.
Active perception provides the alternative framing~\cite{bajcsy1988active}: sensor placement is treated as a decision that resolves a specific uncertainty. Applied to an evolving building, this amounts to asking where to look next in order to determine whether part of the model still describes reality, which requires answering, in sequence, where to observe, how to observe, and when sufficient evidence has been gathered.
This paper takes a first step. While fully realized active perception requires sequential decision-making algorithms and sensor integration, this initial study isolates the geometric prerequisite: evaluating how a robot's embodiment physically constrains the set of available observations before a planner can even choose among them. A mobile manipulator can move the camera viewpoint almost anywhere in its reachable workspace; a fixed chassis sensor cannot. That freedom is commonly assumed valuable and seldom measured. Our contributions are a formulation separating the admissible observation set of an embodiment from the cost of reaching it; a reproducible IFC (Industry Foundation Classes)-derived benchmark that clips a real Revit export into a room-scale scene and injects controlled changes labelled by an observability class describing why they are hard to observe; and an exhaustive sensor-configuration study whose central finding is a distinction we believe generalizes-embodiment partitions scene changes into those a configuration cannot observe at all, and those it can observe only at greater cost.

\section{RELATED WORK}

As-built verification against a building model is well established, from dimensional compliance checking~\cite{bosche2010automated} to progress tracking~\cite{turkan2012automated, golparvar2015automated}. Such systems consume data rather than plan it: coverage or an operator sets the route, and the verification objective does not influence sensor location.

Active perception treats sensing as an action to be planned~\cite{bajcsy1988active}. Next-best-view methods maximize expected information for reconstruction~\cite{connolly1985determination,isler2016information} or exploration~\cite{bircher2016receding}, typically under an unknown-space objective. Our setting differs: a strong prior exists, and the question concerns what has changed. Mounting the sensor on a manipulator extends the achievable viewpoint set, but the magnitude of that extension is rarely quantified.

\section{PROBLEM FORMULATION}

Let $M$ denote prior knowledge of the environment, such as a BIM model, previous map, or registered scan. A sensing configuration $q=(q_b,q_a)$ comprises a mobile-base configuration $q_b$ and an arm configuration $q_a$.

A general active-perception system would select a feasible configuration by trading the expected task-relevant utility of a future observation against its acquisition cost:
\begin{equation}
q^* = \arg\max_{q \in Q} \left( U(q \mid M, B_t) - \lambda C(q) \right)
\label{eq:utility}
\end{equation}
Here $Q$ is the feasible sensing-action space, $B_t$ is the accumulated belief, $U$ is the expected utility of the observation for the current task, $C$ is the cost of acquiring it and the weighting parameter $\lambda \geq 0$ governs the trade-off between information gain and physical effort: a higher $\lambda$ penalizes travel to favor nearby views, while a lower $\lambda$ prioritizes optimal data regardless of distance. The long-term DeltaSeek objective is to instantiate $U$ using evidence such as RGB, depth or point-cloud registration confidence. 

The paper isolates the geometric prerequisite for Eq.~\eqref{eq:utility}. For a scene change or observation target $e$ and sensing configuration $q$, let $V(e, q) \in [0, 1]$ be the visible supporting-surface fraction after range, frustum, incidence, and occlusion tests. An observation is considered geometrically admissible when:
\begin{equation}
 V(e, q) \geq \tau_v, \quad \text{with} \quad \tau_v = 0.05.
\label{eq:visibility-threshold}
\end{equation}

For sensing embodiment $s$ (chassis or wrist), define the admissible observation set:
\begin{equation}
Q_e^s = \{ q \in Q^s : V(e, q) \geq \tau_v \}
\end{equation}

The set $Q_e^s$ provides the central decomposition studied in this paper. If $Q_e^s$ is empty, the target lies outside the sensing capability of embodiment $s$ and no routing policy can recover it. If $Q_e^s$ is nonempty, the observation is feasible and the question becomes how much motion is required to reach one of its admissible configurations. With start configuration $q_0$, an idealized acquisition cost is:
\begin{equation}
C_s(e) = \min_{q \in Q_e^s} d_{\mathrm{nav}}(q_0,q)
\label{eq:acquisition-cost}
\end{equation}
where $d_{\mathrm{nav}}$ is the collision-aware drivable distance. In planned runs, the realized cost is the accumulated route length at which $e$ is first observed.

\section{BENCHMARK}
\subsection{From IFC to a room-scale scene}
The benchmark geometry comes from a Revit 2025 IFC2X3~\cite{liebich1999highlights} export of a single-story building, $44.5 \times 19.0 \times 4.5$~m, comprising 133 products that convert without failure in \texttt{IfcOpenShell}. Each product is approximated by a bounding box oriented in its local placement frame to preserve the source model's $0.90^\circ$ rotation relative to the global axes. The environment was cropped into a single $7.0 \times 12.5$~m room, Hall B [Fig.~\ref{fig:overview}(b)], by cutting each element in its local frame to maintain the correct geometry. Two fit-out units were then added to the scene to serve as physical obstacles for occlusion testing.

\subsection{Observation and Route Models}
Following coverage-based view-planning formulations ~\cite{blaer2009view}, the visibility score $V(e,q)$ in Eq.~\eqref{eq:visibility-threshold} is the fraction of supporting-surface samples that lie within the camera frustum and depth cutoff and satisfy a $75^\circ$ unoccluded incidence limit. For an unmodeled object, the supporting surface is the object itself; for a displaced element, it is the element at its new position or its vacated modeled volume. Both cameras use Intel RealSense D455-equivalent depth intrinsics ($86^\circ\times57^\circ$) and a shared axial-depth cutoff of $r_{\max}=5$~m, below the reported $6$~m maximum~\cite{servi2021metrological}. We set $\tau_v=0.05$ as a permissive evaluation parameter rather than a detector-calibrated threshold. This model assumes perfect recognition/localization and evaluates predefined changes geometrically over $Q^s$, rather than RGB-D detection.

The distance $d_{\mathrm{nav}}$ in Eq.~\eqref{eq:acquisition-cost} is computed using a Dijkstra search over an occupancy grid inflated according to the platform's Nav2 footprint, with a circumscribed radius of $0.606$~m derived from the robot geometry and a grid resolution of $0.06$~m. Accounting for collision-free navigation largely affects the estimated acquisition cost. Across the evaluated viewpoint pairs, the shortest collision-free path is, on average, $1.8\times$ the straight-line distance ($\mathrm{SD}=1.5$, $N=1{,}770$). In the most obstructed case, the required path is $11\times$ the straight-line distance. All conditions start from the admissible grid position nearest the Gazebo spawn, $q_0=(6.796,\,4.681)$~m. A deterministic greedy planner repeatedly selects an affordable, unvisited configuration that maximizes the predicted reduction in unresolved inspection coverage per unit of collision-aware drivable distance. Selected configurations reduce the remaining unresolved fractions according to the nominal visibility model; no sensor observations or posterior updates occur during planning. For each embodiment, the resulting open-loop route has a $25$~m travel budget and is evaluated against all eight injected changes. Cost includes base translation but excludes base rotation, arm motion, and return to the start.
\subsection{Observability Classes}

A change's physical type does not fully characterize its observability. We therefore categorize each change according to its primary source of observational difficulty: \emph{trivial}, a large unmodeled object located in open floor space; \emph{height}, a change located near the ceiling; \emph{incidence}, a change observable only from viewpoints providing a sufficiently favorable surface-incidence angle; and \emph{occluded}, a change hidden by another object from most admissible base poses.

Eight changes were manually placed in Hall B, with two changes assigned to each class. We use a fixed scenario rather than random sampling so that the difficulty of each change can be explained geometrically and the experiment can be reproduced exactly. 
\subsection{Sensing Embodiments}

The platform comprises a Clearpath Husky A300 equipped with a UR5e manipulator. Its robot model is generated from a single \texttt{robot.yaml} configuration, ensuring that the simulation and physical
platform share the same description. The platform uses two identical RGB-D cameras so that differences between the sensing embodiments arise from camera placement rather than sensor capability.

The chassis-mounted camera is fixed to the enclosure deck at $(0.460,\,0.000,\,0.418)$~m in the base frame and pitched downward by $0.17$~rad. Its pose therefore depends only on the base configuration
$q_b$. The wrist-mounted camera uses an eye-in-hand configuration. Its pose is computed through forward kinematics for five predefined arm inspection positions, denoted by $q_a$. The resulting camera poses were verified against the transform tree of the running simulation.

Candidate base positions were generated on a $1.0$~m grid over Hall B. A candidate was retained only if the platform's circumscribed footprint was entirely contained in navigable free space and did not intersect any wall. This filtering yielded 60 base positions. Each retained position was evaluated at four yaw angles, $\theta\in\{0,\pi/2,\pi,3\pi/2\}$, resulting in $\lvert Q_b\rvert=60\times4=240$ admissible base poses.
Combining the 240 base poses with 5 arm postures yields $(240 \times 5=)$ 1,200 distinct wrist-camera poses.
\begin{figure}[t]
    \centering
    \includegraphics[width=\columnwidth]{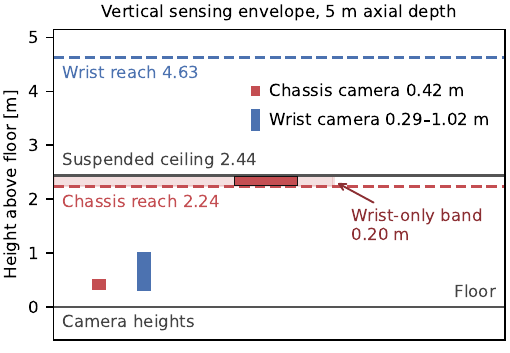}
    \caption{Vertical sensing boundary. At the shared $5$~m axial-depth cutoff, the chassis and wrist cameras reach $2.24$ and $4.63$~m, respectively. The $2.44$~m ceiling leaves a $0.20$~m wrist-only band containing both height
changes.}
    \label{fig:height-bound}
\end{figure}

\begin{figure}[t]
    \centering
    \includegraphics[width=\columnwidth]{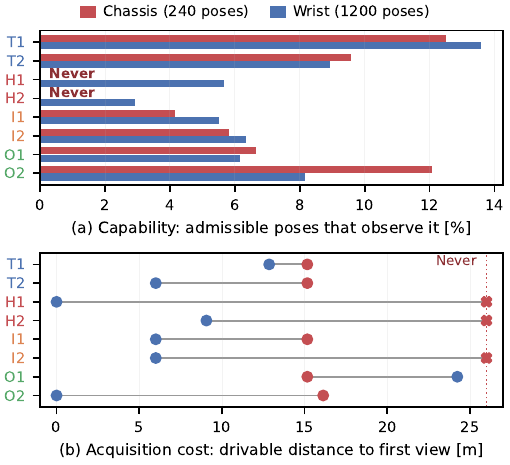}
    \caption{Capability and acquisition cost. (a) Percentage of sensor poses that observe each change. (b) Drivable distance to first observation within the $25$~m multi-target route.}
    \label{fig:results}
\end{figure}
\section{RESULTS}

\subsection{Sensing Capability}

For each sensing embodiment, the planner generates one $25$~m route that is evaluated against all eight changes. The chassis-camera route observes five changes: two of two trivial, zero of two height, one of two incidence, and two of two occluded changes. The wrist-camera route observes all eight changes.

Route-level success alone does not distinguish an embodiment-level capability limitation from an admissible viewpoint that the planner did not visit. We therefore evaluate $V(e,q)$ exhaustively for every $q\in Q^s$. For both height-class changes, $Q_e^{\mathrm{chassis}}=\emptyset$; neither change is visible from any of the 240 chassis-camera poses. These failures therefore reflect a sensing-capability limitation rather than the selected route.

The height failures can be explained geometrically. For a camera mounted at height $h$, pitched downward by $p$, with vertical field of view $v$, the upper frustum boundary at optical-axis depth $r_{\max}$ reaches
\begin{equation}
z_{\max}=h+r_{\max}\left[\cos(p)\tan\left(\frac{v}{2}\right)-\sin(p)\right].
\label{eq:max-visible-height}
\end{equation}

Yaw does not appear in Eq.~\eqref{eq:max-visible-height} because rotation about the vertical axis does not alter the frustum's vertical extent. Applying the equation to the fixed chassis camera and the five wrist
postures gives maximum visible heights of $2.24$ and $4.63$~m, respectively, at the shared $5$~m optical-axis depth cutoff. The latter is the maximum across the evaluated wrist postures. Because the ceiling begins at $2.44$~m, a $0.20$~m wrist-only band remains above the chassis limit. Both height-class changes lie in this band at approximately $2.25$~m. Fig.~\ref{fig:height-bound} illustrates the boundary; the closed-form limits agree with a numerical sweep to within $3$~mm.

The remaining chassis failure is different. The missed incidence-class change is observable from 14 of the 240 chassis poses, but the planner does not visit any of them within its travel budget. It is therefore a routing outcome rather than a physical limitation of the chassis-mounted camera. Fig.~\ref{fig:results}(a) reports the admissible-pose percentages for all eight changes.

\subsection{Acquisition Cost}
The incidence and occlusion cases did not define an absolute sensing-capability boundary. Although these changes are visible from relatively few base poses, at least one admissible chassis viewpoint exists for each. For example, the occluded change behind a pallet stack is visible from only 16 of the 240 chassis-camera poses and is therefore hidden from approximately $93\%$ of cases. Nevertheless, a planner needs to reach only one admissible pose to observe it. For the five changes observed by both embodiments, the median distance traveled before first observation is $6.0$~m along the wrist-camera route and $15.2$~m along the chassis-camera route. Thus, the chassis-mounted camera requires approximately $2.5$ times the median base-translation travel of the wrist camera; arm motion and base rotation are excluded. Thus, occlusion primarily increases acquisition cost rather than making observation physically impossible. Fig.~\ref{fig:results}(b) reports the corresponding distances.

\section{DISCUSSION AND LIMITATIONS}

The results separate sensing capability from acquisition cost. If $Q_e^s=\emptyset$, no route can observe change $e$ using embodiment $s$, and its geometric utility is zero. If $Q_e^s\neq\emptyset$, observation is
possible and the problem becomes the cost $C_s(e)$ of reaching an admissible configuration. This distinction explains why the two height-class failures are embodiment limitations, whereas the missed incidence-class change is a routing outcome. A binary success measure alone would fail to distinguish between these cases.

The capability boundary can also guide hardware design. In Hall B, the $0.20$~m wrist-only band results from the chassis camera's height, downward pitch, field of view, and $5$~m optical-axis depth cutoff. Changing these parameters could reduce or eliminate the band, allowing the framework to identify configuration-specific blind spots before platform deployment.

The study covers only one room, eight injected changes, and geometric visibility with perfect recognition and localization. IFC elements are approximated by boxes, and removing 48 opening entities produces a sealed
room in which the robot is initialized. Observations are evaluated at discrete stations; continuous sensing recovers one incidence-class change but does not alter the height result under the fixed-mount and level-floor
assumptions. Moreover, the height changes exceed the chassis limit by only $11.8$~mm, making this controlled boundary sensitive to camera parameters and the selected depth cutoff rather than a general property of chassis-mounted sensing. Finally, the current planner uses predicted geometric visibility to reduce a heuristic unresolved-uncertainty state, rather than updating a posterior from sensor evidence. Determining when sufficient evidence has been collected will require closed-loop belief updates and an evidence-based stopping rule. Future work will add RGB-D detection, real-robot validation, continuous base-arm planning, randomized multi-room tests, and belief-based stopping.

\section{CONCLUSION}

DeltaSeek presents a foundational evaluation framework for active perception in dynamic environments. We formalized the contribution of the embodiment of a mobile manipulator as the structure of the admissible observation set $Q_e^s$ and the cost $C_s(e)$ of reaching it, and measured both on an IFC-derived benchmark. Two of the eight changes lie entirely outside the chassis camera's admissible observation set. Among the five changes observed by both embodiments, the chassis camera requires approximately 2.5× the median base travel of the wrist camera. A robot that must eventually decide when it has gathered sufficient evidence must first distinguish a view it cannot obtain from the one it has not yet paid to acquire.


\bibliographystyle{IEEEtran}
\bibliography{root}
\end{document}